\documentclass[letterpaper, 10 pt, conference]{ieeeconf}  % Comment this line out if you need a4paper

\IEEEoverridecommandlockouts                              % This command is only needed if 
\usepackage{times}
\usepackage{helvet}
\usepackage{courier}
\usepackage{url}
\usepackage{graphicx}
\usepackage{amsmath} % assumes amsmath package installed
\usepackage{amssymb}  % assumes amsmath package installed
\usepackage{bm}
\usepackage{mathtools}
\usepackage{algorithm}
\usepackage{algpseudocode}
\usepackage[nolist,nohyperlinks]{acronym}
\usepackage{siunitx}
\usepackage{cite}
\usepackage{hyperref}

\newcommand{\deleted}[1]{}
\newcommand{\deletedequation}[2]{}
\newcommand{\revision}[1]{#1}

\usepackage[colorinlistoftodos]{todonotes}

\algnewcommand{\Initialize}[1]{%
  \State \textbf{Initialize:}
  \Statex \hspace*{\algorithmicindent}\parbox[t]{0.96\linewidth}{\raggedright #1}
}

\DeclareMathOperator*{\argmin}{argmin}

\newcommand*{\argminl}{\argmin\limits}

\begin{acronym}
\acro{BO}[BO]{Bayesian Optimization}
\acro{TO}[TO]{trajectory optimization}
\acro{CMAES}[CMA-ES]{Covariance Matrix Adaptation Evolution Strategy}
\acro{RRT}[RRT]{Rapidly-exploring Random Trees}
\acro{GP}[GP]{Gaussian Process}
\acro{EoM}[EoM]{equations of motion}
\acro{NLP}[NLP]{nonlinear program}
\acro{UL}[UL]{upper level}
\acro{LL}[LL]{lower level}
\acro{fSRR}[fSRR]{squared relative residuals}
\acro{CGPUCB}[CGP-UCB]{contextual Gaussian Process upper confidence bound algorithm}
\acro{RMAE}[RMAE]{relative mean absolute error}
\acro{UCB}[UCB]{upper confidence bound}
\acro{base}[B]{base}
\acro{hip}[H]{hip}
\acro{knee}[K]{knee}
\acro{DDP}[DDP]{differential dynamic programming}
\end{acronym}

\graphicspath {{./}}

\title{\LARGE \bf
Locomotion Planning through a Hybrid Bayesian Trajectory Optimization
}

\author{Tim Seyde, Jan Carius, Ruben Grandia, Farbod Farshidian, Marco Hutter% <-this % stops a space
\thanks{Video available at \url{https://youtu.be/zsxcyD60Sjo}}
\thanks{This research was supported by the Intel Network on Intelligent
Systems, the Swiss National Science Foundation through the National Centre of Competence in Research Robotics (NCCR Robotics), the European Union’s Horizon 2020 research and innovation programme under grant agreement No 780883.}% <-this % stops a space
\thanks{All authors are with the Robotic Systems Lab, ETH Zurich, Switzerland
        {\tt\footnotesize \{tseyde,jcarius,rgrandia,farbodf,mahutter\}@ethz.ch} }%
}

\begin{document}

\maketitle
\thispagestyle{empty}
\pagestyle{empty}

%%%%%%%%%%%%%%%%%%%%%%%%%%%%%%%%%%%%%%%%%%%%%%%%%%%%%%%%%%%%%%%%%%%%%%%%%%%%%%%%

\begin{abstract}
Locomotion planning for legged systems requires reasoning about suitable contact schedules. The \revision{contact} sequence and timing\revision{s} \deleted{of contacts defines}\revision{constitute} a hybrid dynamical system and prescribe\deleted{s} a subset of achievable motions. State-of-the-art approaches cast motion planning as an optimal control problem. In order to decrease computational complexity, one common strategy separates footstep planning from\deleted{ the} motion optimization and plans contacts using heuristics. In this paper, we propose to learn contact schedule selection from high-level task descriptors using \acl{BO}. A bi-level optimization is defined in which a \acl{GP} model\deleted{ learns to} predict\revision{s} the performance of trajectories generated by a motion planning \acl{NLP}. \revision{The agent, therefore, retains the ability to reason about suitable contact schedules, while explicit computation of the corresponding gradients is avoided.} We delineate the algorithm in its general form and provide\deleted{ simulation} results for \revision{planning} single-legged hopping. Our method is capable of learning contact schedule transitions that align with human intuition. It\deleted{ furthermore} performs competitively against a heuristic baseline in predicting task appropriate contact schedules. 
\end{abstract}

%%%%%%%%%%%%%%%%%%%%%%%%%%%%%%%%%%%%%%%%%%%%%%%%%%%%%%%%%%%%%%%%%%%%%%%%%%%%%%%%

\section{Introduction}
\revision{Deployment of}\deleted{Introducing} legged systems into real-world scenarios requires autonomous locomotion planning strategies. A challenge in legged locomotion is the necessity to reason about contact schedules during planning. We are interested in the efficient automation of contact schedule selection based on high-level descriptors of a motion task.

The literature provides a wide spectrum of approaches to locomotion planning problems. The separation into simplified models for footstep planning\deleted{ \cite{kajita2001lip,pratt2006capture}} together with robust whole-body control for tracking has found widespread application \revision{\cite{kajita2001lip,pratt2006capture}}. 
In \cite{deits2014footstep}, footstep plans were optimized using a mixed-integer formulation incorporating a reachability heuristic. 
In \cite{mastalli2017trajectory}, the stochastic \ac{CMAES} \cite{hansen2006cma}\deleted{ was used to} optimize\revision{d} footholds\deleted{ from} \revision{on} a pre-labeled terrain costmap. 
Sampling-based methods employing \ac{RRT} have solved complex motion planning problems while being guided by heuristics and requiring post-processing \cite{kuffner2002dynamically,lau2015smooth,bartoszyk2017terrain}.
The strict separation of footstep planning and whole-body tracking requires a conservative approach for the planning of the prior task to ensure feasibility of the latter task. 

In dynamic locomotion tasks, inconsistent contact planning can entail performance degradation or even instability. The motion planning should thus reason about the robot dynamics and the contact states concurrently. 
A continuous-time \ac{DDP} based method was provided in \cite{farshidian17a}, while \cite{pardo2017hybrid,winkler2018gait} proposed direct collocation-based approaches that optimize over the contact schedule while planning dynamically consistent motions. 
These algorithms optimize over the contact duration and can implicitly alter the number of contact phases by setting step durations to zero. 
However, this seems unlikely from a numeric standpoint. 
Furthermore, including the contact schedule in the optimization variables increases \revision{both} the computational burden and the risk to not converge to a feasible solution. 
Similar numerical concerns arise in contact invariant optimization \cite{mordatch2012discovery,posa2014direct}, where contact interaction is implicitly defined by complementary constraints.
This raises the question of whether contact schedule selection can be decoupled from trajectory optimization, without sacrificing model consistency in favor of heuristics.
\begin{figure}[t]
  \centering
  \includegraphics[width=\linewidth]{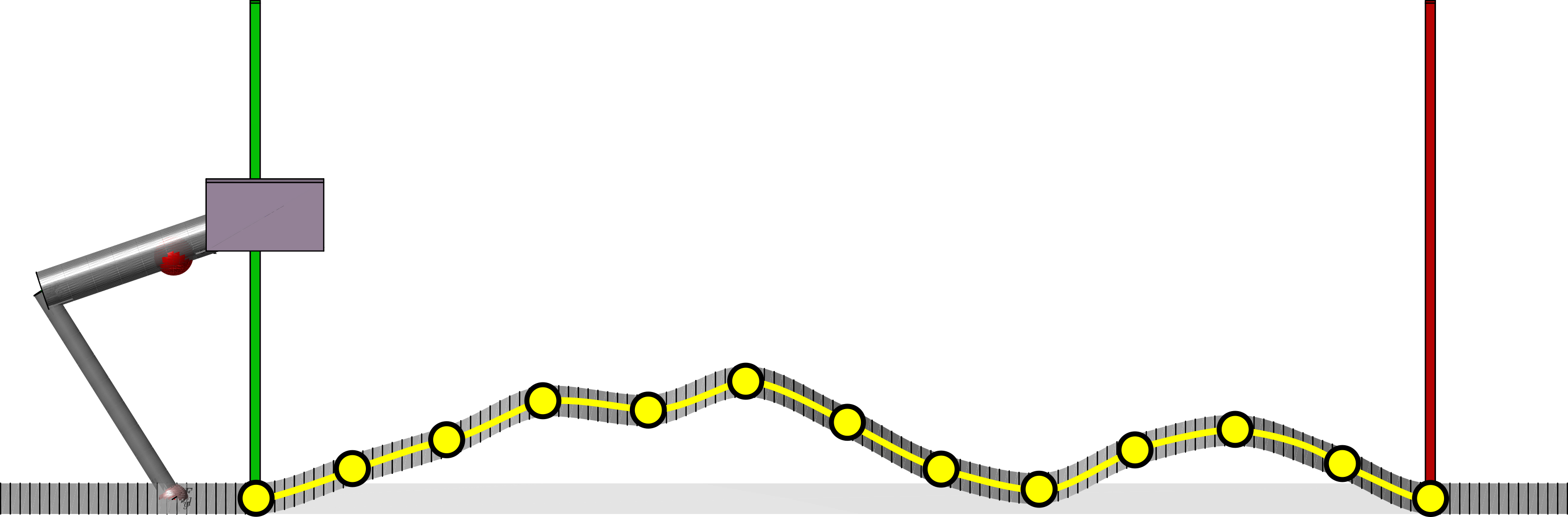}
  \caption{Robot model and environment model. A heightmap is generated by sampling the terrain (yellow dots) and a continuous ground model is recovered via shape-preserving piecewise cubic interpolation (yellow line).}
  \label{fig:Modelling}
\end{figure}

Learning contact schedule selection to guide the trajectory optimization presents itself as a viable option. 
\deleted{Techniques combining learning with optimization to avoid local optima have found widespread interest in the robotics community \cite{farshidian14,levine2014learning,viereck2017learning}.
% describe which tasks and what was done
%
In \cite{hogan2017reactive}, contact mode sequences were learned offline for a manipulation task to simplify the quadratic program solved online.
In \cite{rasmussen2006gaussian,srinivas2009gaussian,krause2011contextualGP}, efficient data acquisition based on \ac{BO} and \ac{GP} models was discussed.
In \cite{rai2017bayesian,calandra2016bayesian}, GPs were applied to automatic controller optimization in locomotion tasks.}
\revision{Techniques combining optimal control with learning have found widespread interest in the robotics community.
Optimization is efficient in finding good local solutions, while data-driven techniques offer the ability to model complex relations. In \cite{farshidian14}, controller adaptation to model inaccuracies was framed as a reinforcement learning problem. In \cite{viereck2017learning} and \cite{levine2013guided}, robustness of neural network policies was demonstrated for locomotion tasks.
In \cite{hogan2017reactive}, a classifier predicts contact mode sequences for a manipulation task under linearized dynamics to simplify the quadratic program solved online.
The strongly nonlinear locomotion tasks considered here make the optimization prone to become locally trapped or to not converge at all. Efficient data acquisition should, therefore, avoid extensive sampling in regions likely to produce infeasible problems.
This exploration-exploitation trade-off can be achieved through a combination of \ac{BO} with \ac{GP} models, as discussed in \cite{rasmussen2006gaussian,srinivas2009gaussian} and applied to contextual bandit problems in \cite{krause2011contextualGP}.
Previously, promising results with these approaches were demonstrated for applications such as automatic locomotion controller optimization in \cite{rai2017bayesian,calandra2016bayesian}.}

In this paper, we propose an efficient method to automate contact schedule selection in optimization-based legged locomotion. The planning problem is cast as a bi-level optimization.
\deleted{Instead of using a gradient-based method \cite{farshidian17b}, or tree search \cite{toussaint2018differentiable}, we explore a learning-based method in the \ac{UL}.
Referring model training to the offline phase effectively reduces computational burden in the online phase.
At the \ac{LL}, a direct collocation method is used to solve the motion planning problem for the state and input trajectories under a given contact schedule. At the \ac{UL}, \ac{BO} is used to optimize a \ac{GP} model which maps the contact scheduling policy to the trajectory optimization performance for a set of tasks. The temporal component of gait planning is therefore decoupled from the continuous-variable optimization (i.e., states and inputs). We finally evaluate the approach on the simulated single-legged hopping robot shown in Fig.~\ref{fig:Modelling}.}
\revision{At the \ac{LL}, a direct collocation method solves the motion planning problem for the state and input trajectories under a fixed contact schedule.
At the \ac{UL}, \ac{BO} is used to optimize a \ac{GP} model which maps the contact scheduling policy to the trajectory optimization performance.
The temporal component of gait planning is therefore decoupled from optimizing continuous state and input trajectories.
Employing a learning-based method in the \ac{UL} allows for sample-efficient offline training that avoids predicted infeasible actions, which can slow down computation significantly \cite{toussaint2018differentiable}. 
Unlike in \cite{farshidian17b}, we furthermore avoid taking gradients with respect to the contact schedule and reduce the risk of getting trapped in local optima.
We evaluate our approach on planning trajectories for the single-legged hopping robot shown in Fig.~\ref{fig:Modelling}.}

\section{Methods}
\label{sec:methods}
Optimization-based locomotion planning attempts to compute optimal state and input trajectories, as well as the underlying contact sequence and durations. This proceeds in accordance with the system dynamics and the environment, as contact imposes constraints on the achievable motions.
In general, operating on these continuous and discrete variables defines a mixed logic optimization problem \cite{bemporad99}, which is computationally expensive to solve. Here, we employ a bi-level optimization\deleted{ method} that separates gait planning (discrete) from trajectory planning (continuous) \cite{farshidian17b}.
At the \ac{LL}, we define a \ac{NLP} to find optimal state and input trajectories under fixed gaits. At the \ac{UL}, we leverage a gradient-free method capable of treating the \ac{NLP} score as a black box function.
The overall problem takes the form
\begin{equation}
\begin{aligned}
    \min_{\mathbf{x}} \mkern-10mu
        & & &
    c_{\text{UL}}\left(\mathbf{x},\hat{\mathbf{y}}\right) 
    \\
    \text{s.t.} \mkern-10mu
        & & &
    \hat{\mathbf{y}} \mkern-5mu = \mkern-5mu \argminl_{\mathbf{y}} 
    \big\{ 
        c_{\text{LL}} (\mathbf{x}, \mathbf{y}) |     
        \mathbf{g}_{\text{LL}} (\mathbf{x}, \mathbf{y}) = \mathbf{0}, \,
        \mathbf{h}_{\text{LL}} (\mathbf{x}, \mathbf{y}) \leq \mathbf{0} 
    \big\},
    \\
        & & &
    \mathbf{g}_{\text{UL}}\left(\mathbf{x}, \mathbf{y}\right) = \mathbf{0}, 
    \\
        & & &
    \mathbf{h}_{\text{UL}}\left(\mathbf{x}, \mathbf{y}\right) \leq \mathbf{0},
    \label{eq:biLevelOpt}
\end{aligned}
\end{equation} 
where $\mathbf{x}$ are the \ac{UL} decision variables,  $\mathbf{y}$ are the \ac{LL} decision variables, $c_{i}$ denotes the cost, $\mathbf{g}_{\text{i}}$ are the equality constraints, and $\mathbf{h}_{\text{i}}$ are the inequality constraints, where ${i \in \{\text{LL}, \text{UL}\}}$.

\subsubsection*{LL optimization}
At the \ac{LL} we consider the gait to be given and set ${\mathbf{x} = \mathbf{x}_{0}}$. We then determine the optimal state and input trajectories for the resulting locomotion task. To this end, we formulate an optimization problem using a direct collocation method \cite{pardo2017hybrid}. The resulting \ac{NLP} is written in its general form as
\begin{equation}
    \begin{aligned}
        \label{eq:NLP}
        \min_{\mathbf{y}} & & & f_{\text{O}}\left(\mathbf{x}_{0},\mathbf{y}\right) \\
        \text{s.t.} & & & \mathbf{g}_{\text{LL}}\left(\mathbf{x}_{0},\mathbf{y}\right) = \mathbf{0}, \\
        & & & \mathbf{h}_{\text{LL}}\left(\mathbf{x}_{0},\mathbf{y}\right) \leq \mathbf{0},
    \end{aligned}
\end{equation}
where ${c_{\text{LL}} = f_{\text{O}}}$ is the \ac{LL} objective function. The \ac{LL} decision variables vector ${\mathbf{y} = \left[\mathbf{y}_{1}^{\top},\dots,\mathbf{y}_{N}^{\top}\right]^{\top}}$ stacks the individual node vectors ${\mathbf{y}_{i} = \left[\mathbf{q}^{\top}_{i},\dot{\mathbf{q}}^{\top}_{i},\mathbf{u}^{\top}_{i}\right]^{\top}}$ consisting of generalized positions $\mathbf{q}$, generalized velocities $\dot{\mathbf{q}}$, and input torques $\mathbf{u}$. The constraints include system dynamics, ground contact, friction, and joint limits. The \ac{NLP} in (\ref{eq:NLP}) returns the optimized trajectory nodes $\mathbf{y}_{\text{opt}}$ together with the associated cost $f_{\text{O,opt}}$ and constraint values $\mathbf{g}_{\text{LL,opt}}$ and $\mathbf{h}_{\text{LL,opt}}$.

\subsubsection*{Merit function}
In order to quantify the quality of solutions provided by the \ac{LL} optimization, we define a merit function $M\left(\mathbf{x},\mathbf{y}\right)$. The merit function weighs trajectory cost and constraint violation based on
\begin{equation}
    \begin{aligned}
    \label{eq:meritFct}
    M\left(\mathbf{x},\mathbf{y}\right) := &\sigma_{1} f_{\text{O}}\left(\mathbf{x},\mathbf{y}\right) + \sigma_{2} \sum_{i=1}^{n} \left(g_{LL,i}\left(\mathbf{x},\mathbf{y}\right)\right)^2\\
    & + \sigma_{3} \sum_{j=1}^{m} \left(\max\left(0,h_{LL,j}\left(\mathbf{x},\mathbf{y}\right)\right)\right)^2,
    \end{aligned}
\end{equation}
where the first term penalizes the trajectory cost, while the second and the third terms penalize constraint violation of the equality and inequality constraints, respectively. \revision{In general, their ratios should ensure a separation between feasible and infeasible runs, while avoiding steep gradients in the merit function.}

\subsubsection*{UL optimization}
At the \ac{UL} we consider the gait to be variable and select $\mathbf{x}$ such that $c_{\text{UL}}$ is minimal. Here, we set  ${c_{\text{UL}} = M\left(\mathbf{x},\mathbf{y}\right)}$ to convey that an optimal gait should lead to both minimal cost and constraint violation. The \ac{UL} optimization is constrained by the \ac{LL} solution such that ${M\left(\mathbf{x}\right) = M\left(\mathbf{x},\mathbf{y} = \text{NLP}\left(\mathbf{x}\right)\right)}$. We therefore choose a gradient-free method from \ac{BO} to solve it. To this end, we model the merit function as a \ac{GP} according to 
\begin{equation}
    \begin{aligned}
        M \left(\mathbf{x}\right) \sim \mathcal{GP} \{\mu \left( \mathbf{x} \right), k\left(\mathbf{x}, \mathbf{x}' \right) \},
    \end{aligned}
\end{equation}
where $\mu \left( \mathbf{x} \right)$ is a mean function and $k\left(\mathbf{x}, \mathbf{x}' \right)$ is a kernel function. The kernel function serves as a similarity measure and relates distance in pairing space ${\mathbf{d}_{\mathbf{x}} = \mathbf{x}-\mathbf{x}'}$ to distance in merit space ${d_{m} = m-m'}$. We denote the set of observed samples by $\{\mathbf{X},\mathbf{m}\}$, where $\mathbf{X}$ is the matrix of observed \ac{UL} decision variables and $\mathbf{m}$ the vector of corresponding merit scores, and a new query by $\{\mathbf{x}^{*},m^{*}\}$. The \ac{GP} model then uses the observations to infer predictions for new queries, while providing estimates of the associated uncertainty. It does so via the predicted mean $\mu$ and predicted standard deviation $\sigma$, defined as
\begin{align}
    \mu\left(\mathbf{x}^{*}\right) &= \mathbf{K}_{\mathbf{x}^{*} \mathbf{X}}\left(\mathbf{K}_{\mathbf{X} \mathbf{X}} + \sigma_{n}^{2} \mathbb{I}\right)^{-1} \mathbf{m}, \\
    \sigma^{2} \left(\mathbf{x}^{*}\right) &= \mathbf{K}_{\mathbf{x}^{*} \mathbf{x}^{*}} - \mathbf{K}_{\mathbf{x}^{*} \mathbf{X}}\left(\mathbf{K}_{\mathbf{X} \mathbf{X}} + \sigma_{n}^{2} \mathbb{I}\right)^{-1} \mathbf{K}_{\mathbf{X} \mathbf{x}^{*}},
\end{align}
where $\mathbf{K}$ is the resulting kernel matrix and $\sigma_{n}$ encodes the noise-level in the observations.

\subsubsection*{Context and actions}
Following the nomenclature in \cite{krause2011contextualGP}, we encode planning tasks using context $\mathbf{z}$. The context here includes the distance to the goal location together with the terrain heightmap. The available gaits are referred to as actions $\mathbf{s}$ which encode a contact sequence and the associated contact durations. We define a context-action pair as ${\mathbf{x} = \mathbf{z} \times \mathbf{s}}$. These parameterize the \ac{UL} optimization and subjected to $\mathbf{z}$ we try to select $\mathbf{s}$ such that $M\left(\mathbf{x}\right)$ is minimal.

\section{Bayesian Optimization Algorithm}
\label{sec:GPregresison}
The concepts introduced in the previous section are combined into a single \ac{BO} algorithm. In the following, we introduce the \ac{GP} regression formulation with the associated termination criterion and further specify both the kernel function and the merit function employed. The corresponding pseudo-code is provided in Algorithm~\ref{alg:regression}.

\subsubsection*{\ac{GP} regression}
We employ \ac{GP} regression to sequentially solve the \ac{UL} optimization in (\ref{eq:biLevelOpt}). Specifically, we use a variation of the \ac{CGPUCB} presented in \cite{krause2011contextualGP}. At iteration $k$, context $\mathbf{z}_{k}$ is sampled uniformly at random from context space $\mathbf{Z}$ and the posterior distribution is evaluated over the entire action space $\mathbf{S}$. Action $\mathbf{s}_{k}$ is then selected using \ac{UCB} sampling according to the acquisition function
\begin{align}
    \mathbf{s}_{k} = \argminl_{\mathbf{s} \in \mathbf{S}} \left(\mu_{k-1}\left(\mathbf{z}_{k},\mathbf{s}\right) - \beta_{k}^{1/2} \sigma_{k-1}\left(\mathbf{z}_{k},\mathbf{s}\right)\right),
\end{align}
where $\mu_{k-1}$ and $\sigma_{k-1}$ denote the posterior mean and standard deviation of the previous iteration, and $\beta_{k} = \log \left(k\right)$ is an exploration-exploitation trade-off variable that encourages ongoing exploration. Based on the resulting $\mathbf{x}_{k}$, we solve the \ac{NLP} and compute a merit score $m_{k}$ from the optimized trajectory. These are then added to the set of observed samples and the \ac{GP} posterior belief is updated. This process continues until a termination criterion is satisfied.

\subsubsection*{Termination criterion}
We assert convergence of the regression by monitoring a discounted version of the relative prediction error. Specifically, we use a first-order lowpass filter on the \ac{fSRR}. At iteration $k$, the expectation of the \ac{fSRR} is computed as
\begin{align}
\mathbb{E} \left[ \left(\frac{\Delta \mathbf{m}}{\mathbf{m}}\right)^{2} \right]_{k} 
  \mkern-10mu &= 
\rho \left(\frac{\Delta m_{k}}{m_{k}}\right)^{2} 
  \mkern-10mu + 
\left(1 - \rho\right) \mathbb{E} \left[ \left(\frac{\Delta \mathbf{m}}{\mathbf{m}}\right)^{2} \right]_{k-1} \mkern-10mu,
\end{align}
where $k$ indicates the current iteration index, $m_{k}$ is the observed merit score, and $\Delta m_{k}$ is the corresponding difference between the observation and the prediction. The scalar $\rho$ defines how much emphasis is placed on the more recent predictions in computing the \ac{fSRR}. Data acquisition is terminated once the \ac{fSRR} falls below a pre-defined threshold provided as
\begin{align}
\mathbb{E} \left[ \left(\frac{\Delta \mathbf{m}}{\mathbf{m}}\right)^{2} \right]_{k}  &\leq \epsilon_{\text{fSRR}}.
\end{align}

\begin{algorithm}[t]
  \caption{Bi-level optimization} \label{alg:regression}
  \begin{algorithmic}[1]
    \Require{$\epsilon_{\text{fSRR}}, \mathbf{Z}, \mathbf{S}, \mathbf{K}_Z, \mathbf{K}_S, \mathbf{\sigma}_Z, \mathbf{\sigma}_S$}
      \Initialize{
      \revision{$\mathbf{X}, \mathbf{m}, \bm{\theta}, \mathbf{K} \gets \{ \}, \{ \}, \left[\bm{\sigma}_{Z},\bm{\sigma}_{S}\right], \mathbf{K}_{Z} \times \mathbf{K}_{S}$}\\
      \deleted{$\hphantom{\epsilon_{\text{fSRR}}}\mathllap{\mathbf{X}} \gets \{ \}$\\
      $\hphantom{\epsilon_{\text{fSRR}}}\mathllap{\mathbf{m}} \gets \{ \}$\\
      $\hphantom{\epsilon_{\text{fSRR}}}\mathllap{\bm{\theta}} \gets \left[\bm{\sigma}_{Z},\bm{\sigma}_{S}\right]$ \\
      $\hphantom{\epsilon_{\text{fSRR}}}\mathllap{\mathbf{K}} \gets \mathbf{K}_{Z} \times \mathbf{K}_{S}$\\}}
    \While{$E_{\text{fSRR}} > \epsilon_{\text{fSRR}}$}
      \State $\mathbf{z}_{k} \gets \Call{Random}{\mathbf{Z}}$
      \item[]
      \State $\beta_{k} \gets \log \left(k\right)$
      \State $\{\mu_{k-1},\sigma_{k-1}\} \gets \Call{Posterior}{\mathbf{X}, \mathbf{m}, \mathbf{K}, \bm{\theta}, \mathbf{z}_{k}, \mathbf{S}}$
      \State $\mathbf{s}_{k} \gets \argminl_{\mathbf{s} \in \mathbf{S}} \left(\mu_{k-1} - \beta_{k}^{1/2} \sigma_{k-1}\right)$  
      \item[]
      \State $m_{k} \gets \Call{NLP}{\mathbf{z}_{k}, \mathbf{s}_{k}}$
      \State $\left(\mathbf{X}, \mathbf{m}\right) \gets \left(\{\mathbf{z}_{k}, \mathbf{s}_{k}\}, m_{k}\right)$
      \item[]
      \State $E_{\text{fSRR},k} \gets \Call{Termination}{\mathbf{m},\bm{\mu}_{k-1},E_{\text{fSRR},k-1}}$
    \EndWhile
  \end{algorithmic}
\end{algorithm}

\subsubsection*{Kernel function}
The \ac{GP} model requires a kernel function $k\left(\mathbf{x},\mathbf{x}'\right)$ as a similarity metric. We follow \cite{krause2011contextualGP} and define one kernel over context space and another kernel over action space. Our overall kernel is then a multiplicative combination of the two, encoding that samples are only similar if both their context and their action are similar:
\begin{align}
    \mathbf{K}\left(\mathbf{x},\mathbf{x}'\right) = \mathbf{K}_{Z}\left(\mathbf{z},\mathbf{z}'\right) \circ \mathbf{K}_{S}\left(\mathbf{s},\mathbf{s}'\right),
\end{align}
with kernel hyperparameters $\mathbf{\sigma}_Z$ and $\mathbf{\sigma}_S$.
For the underlying kernel functions, we choose ARD-Matern3/2 kernels, as these have limited smoothness and therefore have some capability of accommodating potential steps in the sample data. Such discontinuities can arise during transitions from one contact schedule to another, or when one contact schedule suddenly becomes infeasible.

\subsubsection*{Merit refinement}
Failed trajectories can lead to\deleted{ unreasonably} large constraint penalties in the merit function. Such outliers should be avoided, as GP models presume some smoothness characteristics of the functions they attempt to approximate. We therefore subject the merit function $M$ in (\ref{eq:meritFct}) to a sigmoid function to arrive at the new merit function ${M'=\tanh(M)}$. The sigmoid function acts as a soft range limiter and ensures that the merit scores have an upper bound, while the transform is continuous and retains the merit score order.

\section{Modelling}
\label{sec:modelling}
We evaluate the performance of the proposed algorithm on a simulated robot. In the following, the robot model and the environmental model are discussed.

\subsection{System}
The\deleted{ robotic} agent considered is a single-legged hopper, modelled in the sagittal plane (Fig.~\ref{fig:Modelling}). It consists of a \ac{base},\deleted{ a} thigh,\deleted{ a} shank and actuated joints at the \ac{hip} and \ac{knee}. The generalized coordinates $\mathbf{q}$ and joint torques $\mathbf{u}$ are
\begin{align}
\mathbf{q} &= \left[x_{\text{B}}, z_{\text{B}}, \varphi_{\text{H}}, \varphi_{\text{K}}\right]^{\top}, \\
\mathbf{u} &= \left[u_{\text{H}}, u_{\text{K}}\right]^{\top},
\end{align} 
where $x_{i}$ is a horizontal position, $z_{i}$ is a vertical position, $\varphi_{i}$ is a joint angle, and $u_{i}$ is the corresponding joint torque.
The associated \ac{EoM} are written as
\begin{equation}
\label{eq:eom}
\mathbf{M}\left(\mathbf{q}\right) \ddot{\mathbf{q}} + \mathbf{b}\left(\mathbf{q},\dot{\mathbf{q}}\right) + \mathbf{g}\left(\mathbf{q}\right) = \mathbf{S}^{\top} \mathbf{u} + \mathbf{J}_{\text{c}}^{\top} \bm{\lambda}_{\text{c}},
\end{equation} 
where ${\mathbf{M} \in \mathbb{R}^{4 \times 4}}$ denotes the inertia matrix, ${\mathbf{b} \in \mathbb{R}^{4}}$ groups Coriolis and centrifugal effects, ${\mathbf{g} \in \mathbb{R}^{4}}$ is the gravitational force vector, ${\mathbf{S} \in \mathbb{R}^{2 \times 4}}$ is the selection matrix, ${\bm{\lambda}_{\text{c}} \in \mathbb{R}^{2}}$ refers to the ground reaction force, and ${\mathbf{J}_{\text{c}} \in \mathbb{R}^{2 \times 4}}$ is the corresponding Jacobian. 

The system dynamics are represented as a hybrid system by projecting the rigid body equations into the null space of the contact constraints. During flight, no external force is present and equation~\eqref{eq:eom} can be solved to obtain the generalized accelerations. During stance, friction limited ground contact occurs at the point foot. The associated contact constraint is derived to yield
\begin{equation}
\label{eq:contact_constraint_acceleration}
\mathbf{J}_{\text{c}}\ddot{\mathbf{q}} + \dot{\mathbf{J}}_{\text{c}} \dot{\mathbf{q}} = \mathbf{0},
\end{equation}
which is used to solve \eqref{eq:eom} simultaneously for the accelerations and contact force. The stance dynamics are\deleted{ taken from}
\begin{equation}
\begin{bmatrix}
  \mathbf{M} & -\mathbf{J}_{\text{c}}^{\top} \\
  -\mathbf{J}_{\text{c}} & \mathbf{0}
\end{bmatrix}
\begin{bmatrix}
  \ddot{\mathbf{q}} \\
  \bm{\lambda}_{\text{c}}
\end{bmatrix} = 
\begin{bmatrix}
  \mathbf{S}^{\top} \mathbf{u} - \mathbf{b} - \mathbf{g} \\
  \dot{\mathbf{J}}_{\text{c}} \dot{\mathbf{q}}
\end{bmatrix}.
\end{equation}
The resulting contact forces are furthermore constrained in our \ac{TO} to lie in the friction cone defined by the local terrain normal.

\subsection{Environment}
The environment is encoded via a heightmap representation. Terrain height $h$ is sampled along the horizontal with discretization $\Delta x$ on the interval ${x \in [x_{\text{min}}, x_{\text{max}}]}$. For a total of $n_{\text{S}}$ samples, the resulting heightmap is
\begin{align}
\begin{split}
\mathbf{x}_{\text{hm}} = \left[\phantom{h(}x_{\text{hm,1}}\phantom{)}, \ldots, \phantom{h(}x_{\text{hm},n_{\text{S}}}\phantom{)}\right], \\
\mathbf{z}_{\text{hm}} = \left[h(x_{\text{hm,1}}), \ldots, h(x_{\text{hm},n_{\text{S}}})\right].
\end{split}
\end{align}
A $C^{1}$- continuous representation of the terrain is recovered by leveraging shape-preserving piecewise cubic interpolation. Terrain height and gradient at intermediate locations are then computed from the interpolation polynomials.

\section{Experiments}
\label{sec:experiments}
Our main interest is in whether the \ac{GP} model can learn reasonable contact schedule transitions for a given task. We are furthermore interested in whether the resulting predictor can perform competitively against a heuristic baseline. In the following, we introduce the general experimental setup and provide results for planning motions on both flat ground and uneven terrain.

\subsection{Preliminaries}
The following describes the general setup used for both the flat ground scenario and the uneven terrain scenario. First, the representation of context and action is provided. Then, the algorithm used for assessing model performance is outlined.

\subsubsection{Context-action space}
The input space of the regression is spanned by the context space $\mathbf{Z}$ and the action space $\mathbf{S}$. A context ${\mathbf{z} \in \mathbf{Z}}$ consists of a\deleted{ single} goal distance and a\revision{n} $n_{\text{T}}$-dimensional vector of terrain features, denoting\deleted{ the} varying heightmap values. We consider goal distances of up to ${z_{\text{D}} = 1.0\si{m}}$ and terrain height magnitudes of up to ${|z_{\text{T}}| = 0.2\si{m}}$, such that
\begin{align}
\mathbf{Z} &= \left[0.0\si{m},1.0\si{m}\right] \times \left[-0.2\si{m}, 0.2\si{m}\right]^{n_{\text{T}}}.
\end{align}
An action ${\mathbf{s} \in \mathbf{S}}$ consists of an $n_{\text{p}}$-dimensional vector of alternating contact/flight phase durations. As we employ collocation nodes with constant time length, phase durations are encoded by the number of nodes.  We consider up to two jumps, ${n_{\text{p}} = 5}$, and phase node numbers in the range of ${n_{\text{N}} = \left[3,6\right]}$, as these encourage dynamic motions with lower jerk. We then have
\begin{align}
\mathbf{S} &= \{3,4,5,6\} \times \{0,3,4,5,6\}^{n_{\text{p}}-1}.
\end{align}
The \ac{UL} optimization determines the desired gait by setting certain phases to zero. For example, a single stance phase is encoded by action ${\mathbf{s}_{\text{A}} = \left[n_{1},0,0,0,0\right]}$, a single jump by ${\mathbf{s}_{\text{B}} = \left[n_{1},n_{2},n_{3},0,0\right]}$, and a double jump by ${\mathbf{s}_{\text{C}} = \left[n_{1},n_{2},n_{3},n_{4},n_{5}\right]}$. Our nomenclature requires ${\mathbf{s}_{\text{D}} = \left[n_{1},0,0,n_{4},n_{5}\right]}$ to be encoded by ${\mathbf{s}_{\text{B}}}$.

\subsubsection{Performance metric}
Performance of a trained GP model is assessed on a set of ${n_{\mathbf{z}}}$ contexts sampled uniformly at random. For each context, the model predicts an optimal action and the merit score is determined by running the optimization. The procedure is outlined in Algorithm~\ref{alg:prediction}.
%The baseline also selects an action and receives its resulting merit score. The procedure is outlined in Algorithm~\ref{alg:prediction}.
\begin{algorithm}[t]
  \caption{Performance evaluation} \label{alg:prediction}
  \begin{algorithmic}[1]
      \Initialize{\revision{
      $\text{Player1}, \text{Player2} \gets \{\mathbf{X}, \mathbf{y}, \mathbf{K}, \bm{\theta}, \mathbf{S}\}, \mathbf{S}_{\text{Baseline}}$ \\
      $\hphantom{\text{Player1}, \text{Player2}}\mathllap{k_{\text{max}}, n_{\text{Wins}}} \gets 100$, \{0, 0\}}
      \deleted{$\hphantom{\text{Player1}}\mathllap{k_{\text{max}}} \gets 100$\\
      $\hphantom{\text{Player1}}\mathllap{n_{\text{Wins}}} \gets \{0, 0\}$ \\
      $\text{Player1} \gets \{\mathbf{X}, \mathbf{y}, \mathbf{K}, \bm{\theta}, \mathbf{S}\}$\\
      $\text{Player2} \gets \mathbf{S}_{\text{Baseline}}$\\}}    
      \For{$k = 1$ to $k_{max}$}
      \State $\mathbf{z}_{k} \gets \Call{Random}{\mathbf{Z}}$
      \item[]
      \State $\{\mu_{k-1},\sigma_{k-1}\} \gets \Call{Posterior}{\mathbf{X}, \mathbf{y}, \mathbf{K}, \bm{\theta}, \mathbf{z}_{k}, \mathbf{S}}$
      \State $\mathbf{s}_{k,\text{Player1}} \gets \argminl_{\mathbf{s} \in \mathbf{S}} \mu\left(\mathbf{s},\mathbf{z}_{k}\right)$
      \State $m_{k,\text{Player1}} \gets \Call{Optimization}{\mathbf{z}_{k}, \mathbf{s}_{k,\text{Player1}}}$
      \item[]
      \State $\mathbf{s}_{k,\text{Player2}} \gets \Call{Baseline}{\mathbf{z}_{k}, \mathbf{S}_{\text{Baseline}}}$
      \State $m_{k,\text{Player2}} \gets \Call{Optimization}{\mathbf{z}_{k}, \mathbf{s}_{k,\text{Player2}}}$
      \item[]
      \State $n_{\text{Wins}} \gets \Call{UpdateScores}{m_{k,\text{Player1}},m_{k,\text{Player2}}}$
    \EndFor
  \end{algorithmic}
\end{algorithm}

\subsection{Scenario 1: flat ground}
The scenario of hopping over flat ground is considered first. We refer to this flat terrain as FT with ${n_{\text{T}}=0}$, i.e., the terrain is not part of the context for this experiment.

\subsubsection{Learning contact transitions}
\begin{figure}[t]
  \centering
  \includegraphics[width=\linewidth]{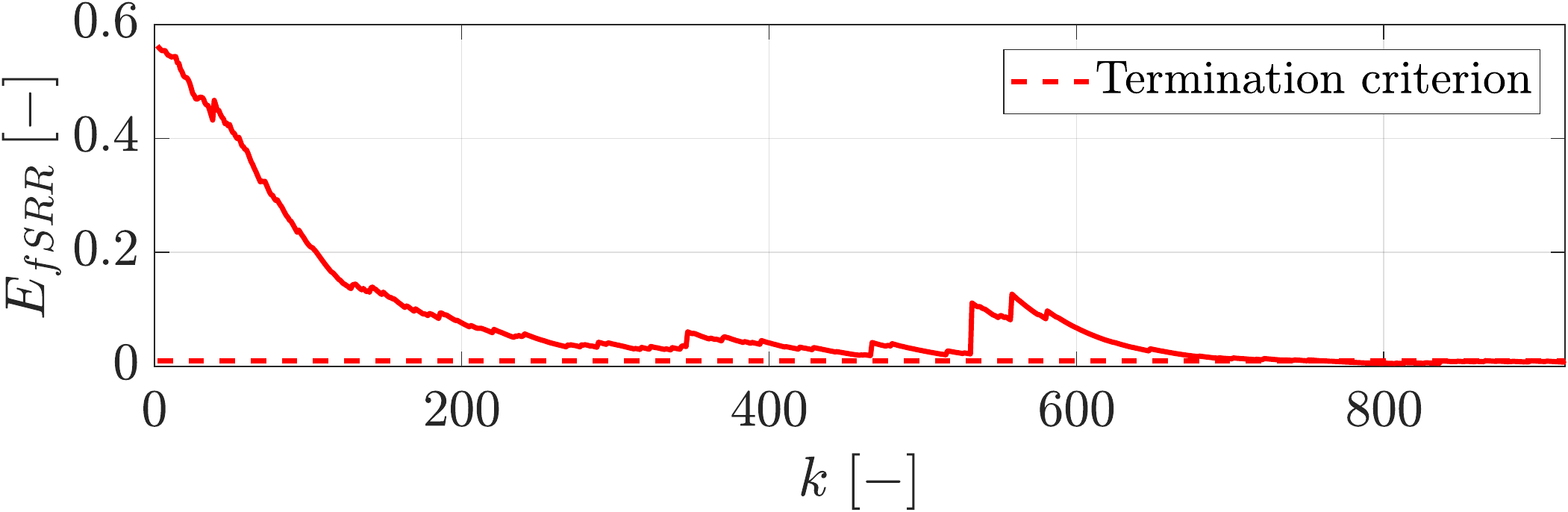}
  \caption{Termination criterion based on \ac{fSRR} over iterations. Intermittent increases in the \ac{fSRR} correlate with the severity of posterior updates, i.e., significant differences between prediction and observation.}
  \label{fig:MeritPredictionGP0}
\end{figure}

\begin{figure}[t]
  \centering
  \includegraphics[width=\linewidth]{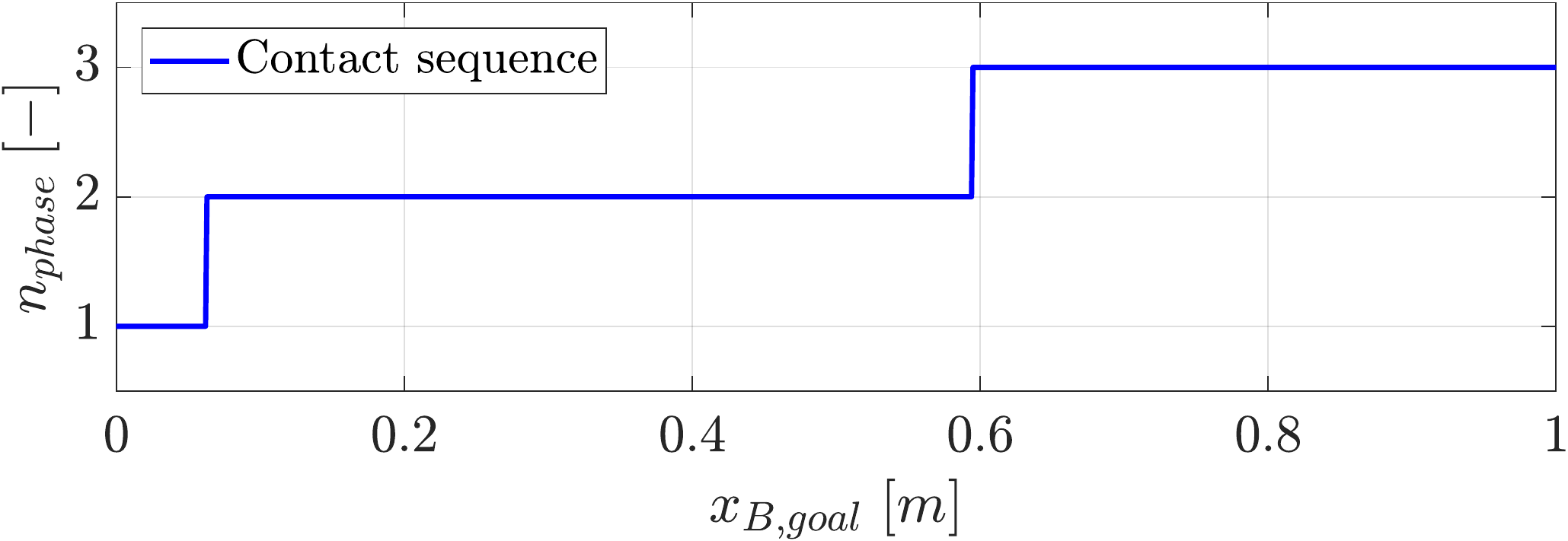}
  \caption{Learned contact sequence transitions over goal distances (resolution ${\Delta x = 0.001\si{m}}$). A natural transition from stance over single jump to double jump emerges.}
  \label{fig:ContactScheduleGP0}
\end{figure}

\begin{figure}[t]
  \centering
  \includegraphics[width=\linewidth]{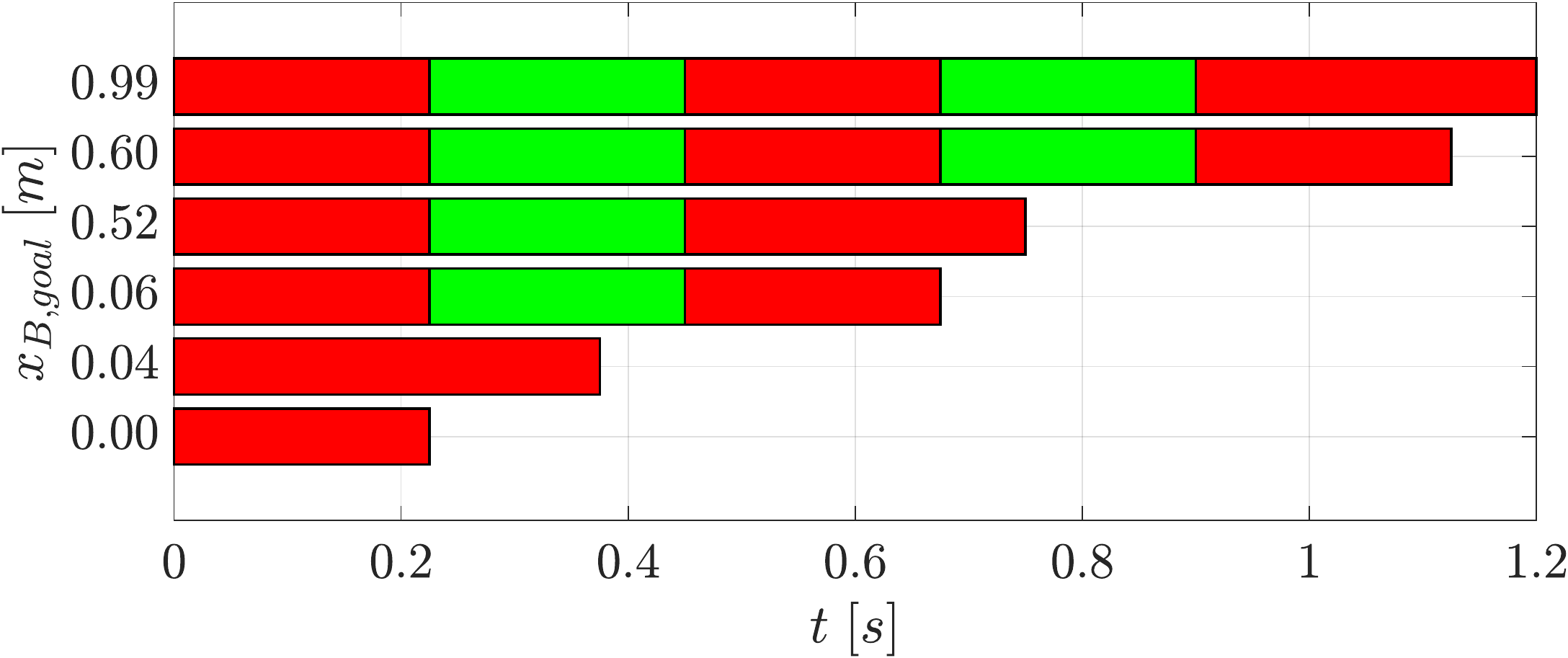}
  \caption{Learned contact schedules consisting of stance (red) and flight (green) phases at specific goal distances. For each contact sequence, two schedules are displayed with the goal distance at which they occur first.}
  \label{fig:ContactDurations_GP0v84}
\end{figure}

We trained a \ac{GP} model on flat terrain via Algorithm~\ref{alg:regression} and refer to it as \ac{GP}-FT. Fig.~\ref{fig:MeritPredictionGP0} provides the \ac{fSRR} termination criterion over the number of iterations. It is apparent that the error metric is steadily decreasing, which means that the predicted merit scores become more accurate over time. Discontinuities in the \ac{fSRR} indicate larger updates in the posterior belief, which can happen when exploring new parts of the context-action space. Fig.~\ref{fig:ContactScheduleGP0} provides an overview of the learned contact sequence transitions over different goal distances, while Fig.~\ref{fig:ContactDurations_GP0v84} displays a selection of learned contact schedules. From Fig.~\ref{fig:ContactScheduleGP0} it is evident that the \ac{GP} model is capable of learning natural contact sequence transitions that progress from stance over a single jump to a double jump. Referring to the contact schedules in Fig.~\ref{fig:ContactDurations_GP0v84}, the model furthermore learned to vary contact durations within the contact sequences. The overall trend of adapting the phase durations or adding steps to the contact schedule as goal distance increases matches what human intuition would suggest.

\subsubsection{Performance}
\begin{figure}[t]
  \centering
  \includegraphics[width=\linewidth]{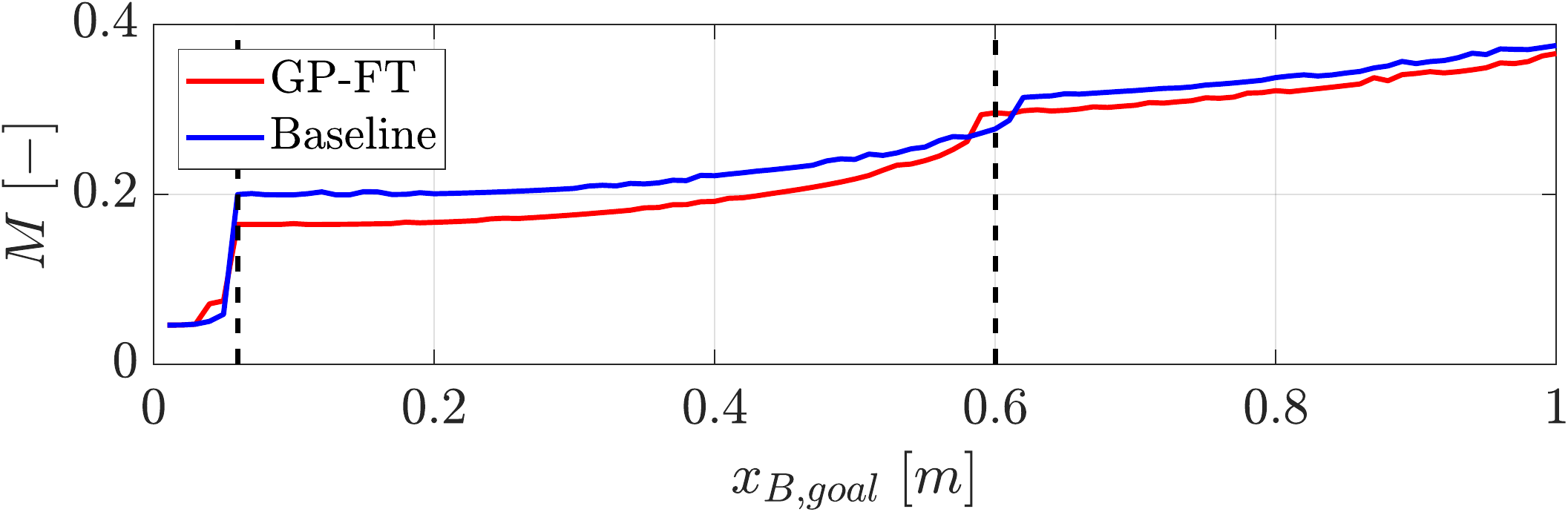}
  \caption{Merit scores for flat ground hopping: GP-FT model and baseline. Dashed lines mark contact sequence transitions in the trained model. The model outperforms the baseline.}
  \label{fig:PredictionGP0Baseline}
\end{figure}

We compare the performance of our learned model to a heuristic baseline. The baseline runs the trajectory optimization for a fixed set of $5$ contact schedules and selects the best action based on the resulting merit scores. The individual schedules were selected before model training based on intuition. Initial and final stance phases were chosen sufficiently long to initiate and terminate dynamic hopping. Intermediate stance phases are shorter to keep motions fluid. Flight phases are short to limit energy expenditure. The action-set available to the baseline is
\begin{align}
\mathbf{S}_{\text{Baseline}} =
\begin{cases}  
\left[3,0,0,0,0\right],\\
\left[4,3,5,0,0\right],\\
\left[5,4,6,0,0\right],\\
\left[4,3,3,3,4\right],\\
\left[5,4,3,4,6\right].
\end{cases}
\end{align}
The performance of the GP-FT  model was evaluated by competing against the baseline according to Algorithm~\ref{alg:prediction} over ${n_{\mathbf{z}} = 100}$ runs. The results are provided in Fig.~\ref{fig:PredictionGP0Baseline} which illustrates the merit scores of different goal distances for the baseline and GP-FT approaches. In general, the trained GP-FT has a better performance than the baseline method. At the phase transitions, performance seems to converge with the baseline having a slight edge at the stance to single jump and single jump to double jump transition. 
%Overall, the trained GP-FT model learned to select optimal actions based on the provided context.

\subsection{Scenario 2: uneven terrain}
\begin{figure}[t]
  \centering
  \includegraphics[width=\linewidth, trim={0 0 0 0.0cm},clip]{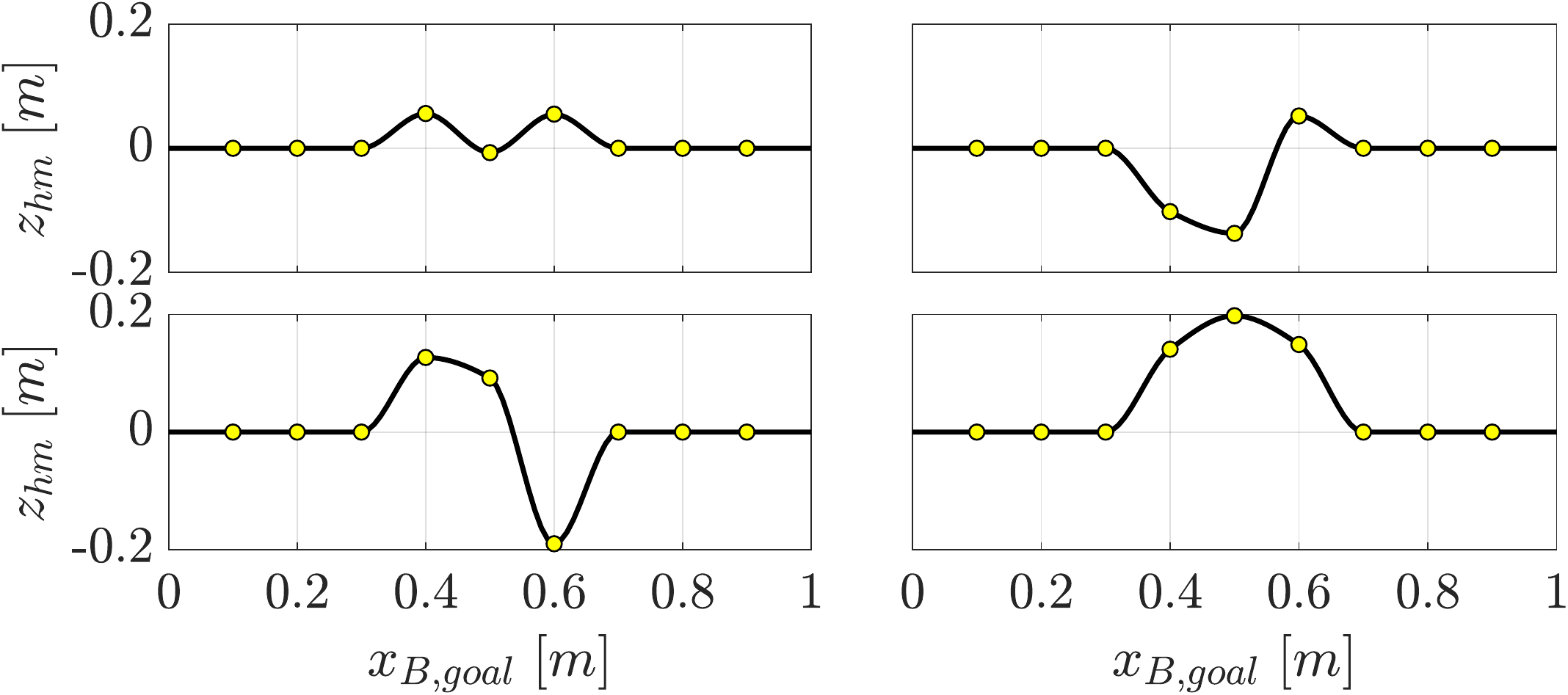}
  \caption{Sample terrains: three of the underlying terrain nodes (yellow) are randomly sampled within bounds.}
  \label{fig:SampleTerrains}
\end{figure}

The scenario of hopping over uneven terrain is considered next. Fig.~\ref{fig:SampleTerrains} shows some of the randomly generated terrains used for training of the GP.  
%and each instance is referred to as Terrain R. 
Here, we vary $3$ terrain nodes to create obstacles. Thus, in the following, we use ${n_{\text{T}}=3}$ features as terrain context to describe the heightmap.

\subsubsection{Learning contact transitions}
We trained a GP model on randomly generated rough terrain via Algorithm~\ref{alg:regression} and refer to it as GP-RT. This proceeded in two stages. First, the model was trained on flat ground until convergence to an intermediate \ac{fSRR} value. This initializes the model with unperturbed samples from the underlying system dynamics. Then, the model was trained on rough terrain until convergence to a final \ac{fSRR} value. At each iteration, a terrain was randomly sampled from a Gaussian distribution. 

\subsubsection{Performance}
\begin{figure}[t]
  \centering
  \includegraphics[width=\linewidth, trim={0 0 0 0.0cm},clip]{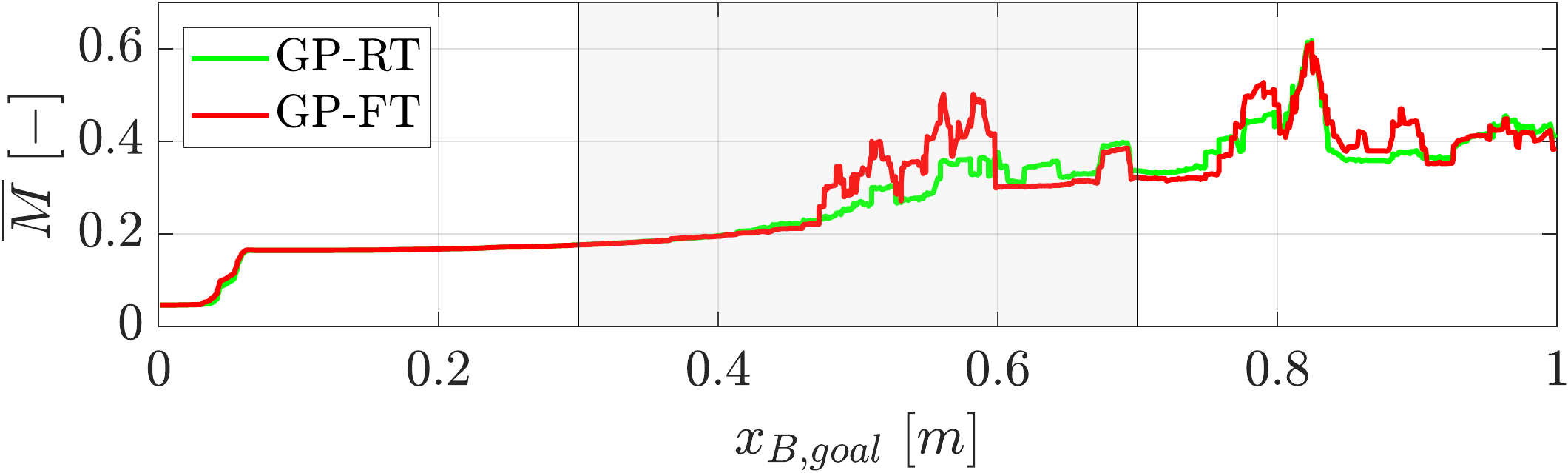}
\caption{Performance comparison of the GP-RT and GP-FT models over rough terrain.  Merit scores were obtained over ${n_{\mathbf{z}} = 1000}$ random contexts and are displayed in their filtered form (moving average). The GP-RT model overall provides actions incurring lower merit scores.}
  \label{fig:PredictionGPRGP0}
\end{figure}

\begin{figure}[t]
  \centering
\includegraphics[width=0.992\linewidth, trim={0 0 0 0.0cm},clip]{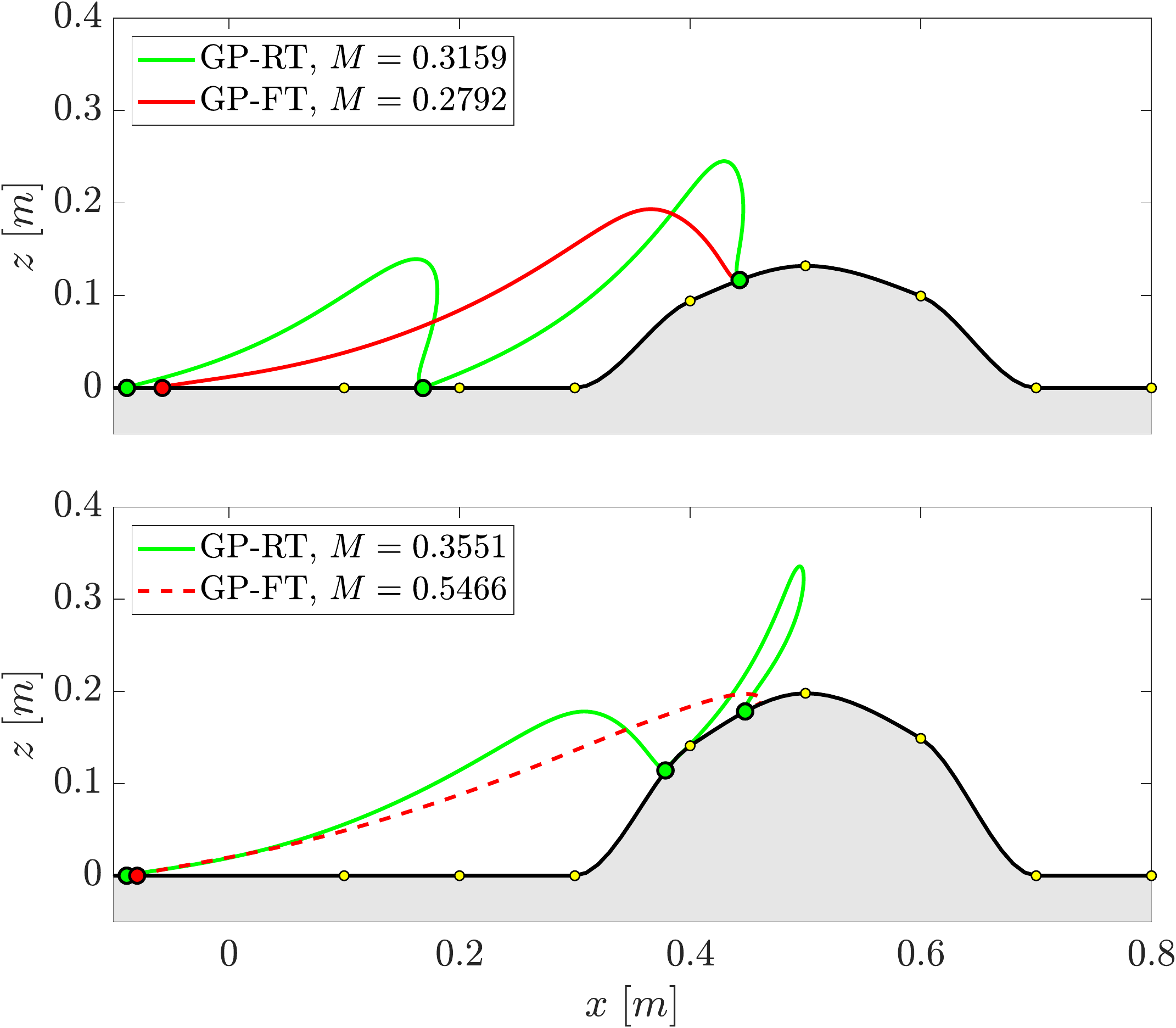}
  \caption{Foot trajectories of the GP-RT and GP-FT models for hopping onto obstacles. On both, the GP-RT model decides to employ an additional step. While GP-RT performs slightly worse than GP-FT on the small obstacle, it clearly outperforms GP-FT on the bigger obstacle (dotted line marks premature termination).}
  \label{fig:Footholds}
\end{figure}

The performance of the GP-RT model was evaluated by competing against the GP-FT model according to Algorithm~\ref{alg:prediction} over ${n_{\mathbf{z}} = 1000}$ runs. The results are provided in Fig.~\ref{fig:PredictionGPRGP0}. The merit scores\deleted{ are} \revision{were} smoothed\deleted{ out} using a moving average filter for\deleted{ interpretation purposes} \revision{interpretability}.
The GP-RT model incurs lower merit scores than the GP-FT model on average. This can\deleted{ largely} \revision{in part} be attributed to the GP-RT model selecting actions that fail less frequently than those selected by the GP-FT model. 
Here, GP-FT generated motions fail in about $6\%$ of the evaluated runs, while GP-RT motions fail only in $3\%$ of them. Note that \revision{we restrict our analysis to a maximum of two jumps and randomly sampled context may generate truly infeasible objectives (e.g., Fig.~\ref{fig:Footholds} bottom with the maximum goal distance ${z_{\text{D}} = 1.0\si{m}}$).}\deleted{ heightmaps are randomly generated, and some truly infeasible scenarios should be expected.} 

In general, the GP-FT model acts aggressively and selects very dynamic motions, or short contact schedules, at the risk of failing. \deleted{On the other hand, t}\revision{T}he GP-RT model acts more conservatively and adapts its actions to the observed terrain to avoid failing. For this reason, we observed\deleted{ that} the GP-RT actions\deleted{ tend} to perform slightly worse than the GP-FT actions on terrains with limited roughness. However, GP-RT motions demonstrate an increased level of robustness on terrains with more severe roughness. This behavior can be attributed to the underlying probabilistic model used for modeling the context (i.e., terrain heightmap). In the GP-RT case, the optimized gait policy always assumes some level of uncertainty over the observed terrain. To avoid failure the GP-RT policy chooses gait sequences which are less prone to uncertainties such as premature \ac{TO} convergence or falling into a local minimum. 

An example of this robust behavior is provided in Fig.~\ref{fig:Footholds}. Here, the GP-RT model selects a double jump schedule, while the GP-FT model selects a single jump schedule. On the smaller obstacle (top), the GP-RT performs slightly worse than the GP-FT. On the higher obstacle (bottom), the GP-RT performs significantly better than the GP-FT. This is because the lower level NLP for the latter does not converge. Learning the terrain features is therefore advantageous in selecting contact schedules competitively. 

\section{Conclusion}
\label{sec:conclusion}
In this paper, we proposed a method that learns to select contact schedules based on high-level task descriptors. The problem is cast as a bi-level optimization, where contact planning proceeds in the \ac{UL} and constrained \ac{TO} in the \ac{LL}. The performance of a trajectory resulting from a specific contact schedule task is quantified using a merit function. The merit function is modeled as a \ac{GP} and contact schedule selection is learned via a \ac{BO} approach. 

It was shown that the \ac{GP} model is capable of learning contact schedule transitions that\deleted{ correspond to what} \revision{align with} human intuition\deleted{ would predict}. \deleted{It was furthermore shown that t}\revision{T}he trained model is capable of outperforming a heuristic baseline in predicting task appropriate contact schedules. During training, the \ac{GP} model learns about both the underlying system dynamics and the interaction with the specific terrain. \revision{It hereby does not necessarily find the global optimum of the strongly non-convex problem, but learns to select contact schedules that the \ac{NLP} solver performs well with.} 

It was demonstrated that\deleted{ the performance of} a model trained on rough terrain outperforms a model trained on only flat terrain,\deleted{ showing} \revision{highlighting} that the method can incorporate terrain information into the decisions. However, we were impressed by how close the model trained\deleted{ only} on flat terrain comes to the rough terrain performance. It shows that our bi-level formulation provides \revision{a} certain robustness. Even when the \ac{UL} does not modify its decision to the terrain, the \ac{LL} optimizes with full terrain information and manages to find reasonable solutions. The main difference is seen in scenarios with extreme terrain features, where the \ac{TO} cannot converge\deleted{ to a solution} with the flat terrain gait while the rough terrain gait succeeds in the task.

Future work will include the extension to longer gait sequences over more complex terrains. Moreover, we wish to use the method in a 3D setting as well. For both aspects, the scalability of the \ac{GP}s to the larger input space has to be addressed\revision{, as exact \ac{GP} regression has ${\mathcal{O}\left(n^3\right)}$ complexity}. A promising strategy would be to employ a Deep Neural Network to imitate the \ac{GP} model and thus to achieve better scalability at the expense of longer training sessions.

%%%%%%%%%%%%%%%%%%%%%%%%%%%%%%%%%%%%%%%%%%%%%%%%%%%%%%%%%%%%%%%%%%%%%%%%%%%%%%%%

\addtolength{\textheight}{-8.7cm}
\bibliographystyle{./IEEEtran}
\bibliography{./bayes_traj_opt.bib}

\end{document}